\pdfoutput=1

\documentclass[11pt]{article}

\usepackage[]{ACL2023}

\usepackage{times}
\usepackage{latexsym}

\usepackage[T1]{fontenc}

\usepackage[utf8]{inputenc}

\usepackage{microtype}

\usepackage{inconsolata}

\usepackage{amsmath}

\usepackage{bbm}

\usepackage{booktabs}
\usepackage{graphicx}

\title{The Chai Platform's AI Safety Framework}

\author{Xiaoding Lu \And Aleksey Korshuk \And Zongyi Liu \And William Beauchamp\\
  \AND Chai Research}

\begin{document}
\maketitle
\begin{abstract}
Chai empowers users to create and interact with customized chatbots, offering unique and engaging experiences. Despite the exciting prospects, the work recognizes the inherent challenges of a commitment to modern safety standards. Therefore, this paper presents the integrated AI safety principles into Chai to prioritize user safety, data protection, and ethical technology use. The paper specifically explores the multidimensional domain of AI safety research, demonstrating its application in Chai's conversational chatbot platform. It presents Chai's AI safety principles, informed by well-established AI research centres and adapted for chat AI. This work proposes the following safety framework: Content Safeguarding; Stability and Robustness; and Operational Transparency and Traceability. The subsequent implementation of these principles is outlined, followed by an experimental analysis of Chai's AI safety framework's real-world impact. We emphasise the significance of conscientious application of AI safety principles and robust safety measures. The successful implementation of the safe AI framework in Chai indicates the practicality of mitigating potential risks for responsible and ethical use of AI technologies. The ultimate vision is a transformative AI tool fostering progress and innovation while prioritizing user safety and ethical standards.
\end{abstract}

\section{Introduction}
With the rapid improvement in the quality and fluency of virtual conversational AI agents, there has been growing integration of chat-based based AI systems (Chat AIs) into a range of real-world applications~\cite{DBLP:journals/corr/abs-2201-06657, chatbotsurvey}. This has led to the development of new technologies and opportunities, as seen in platforms such as the Chai research platform. The Chai research platform is an innovative online platform that empowers users to design and interact with chatbots that emulate friends, mentors, or fictional characters~\cite{irvine2023rewarding}. 

However, with this flexibility comes potential challenges. If unregulated, there's a risk that users, intentionally or unintentionally, might design chatbots that do not align to desired ethical and safe AI standards. This potential issue highlights the importance of robust safety measures to prevent harmful outcomes, such as promoting inappropriate content or negatively impacting user well-being~\cite{lu2023safer}. This paper addresses safety considerations for Chat AI and introduces the AI safety principles integrated into the Chai platform to ensure that deployed chat ai technologies best align with modern safe AI practices~\cite{schuett2023best}. Safe AI is about developing and managing AI systems that work in a manner that is beneficial to humanity and encourages generation of safe content. It involves prioritising user safety, protecting data, and ensuring that the technology behaves ethically and responsibly~\cite{10.1145/3551385}. In this paper, we detail our three main chat ai safety pillars; content safeguarding~\cite{AwadDsouzaKimSchulzHenrichShariffBonnefonRahwan2018}, system stability and robustness~\cite{DBLP:journals/corr/abs-2112-00639}, and operational transparency and traceability~\cite{WannerHermHeinrichJaniesch2022}. The second part of this paper then discusses how these considerations are practically integrated into chat ai platforms, where we detail the strategies taken that reinforces Chai's commitment to creating a safe AI conversational platform.

\section{Background}

AI safety, as a field of study, is dedicated to understanding and mitigating the potential risks associated with the development and deployment of artificial intelligence technologies~\cite{bostrom2014superintelligence, schuett2023best}. The primary concern revolves around ensuring alignment~\cite{Christian_2020, DBLP:journals/corr/abs-2101-06060} of these technologies with human values, ethics, and societal norms~\cite{Yudkowsky2011ComplexVS}. This involves designing AI systems that can understand, learn from, and act in accordance with the principles, values, and goals that humans hold~\cite{DBLP:journals/corr/abs-2001-09768, Sotala_2015}. The objective is to create artificial intelligence that not only improves efficiency and productivity but also respects human autonomy, privacy, and other core ethical principles~\cite{Dignum_2019}.

In order to achieve this, AI safety research~\cite{10.1145/3551385} is multidimensional, involving elements of computer science, cognitive science, ethics, and social science~\cite{Fishbein_Ajzen_2005, Sarma_Hay_2017}. Researchers investigate technical aspects like robustness, interpretability, and transparency of AI systems, while also exploring philosophical questions about moral and ethical considerations regarding how we define human values~\cite{TurchinManuscript-TURAAP, Muehlhauser_Helm_2012, Friedman_Hendry_2019}. A range of approaches explore how to effectively incorporate defined human values within reward functions used to train AI systems~\cite{DBLP:journals/corr/abs-2008-02275, DBLP:conf/aaai/SoaresFAY15, 10.5555/645529.657801, Riedl2016UsingST, StuartArmstrong, 10.1145/2955091}. The ultimate aim is to ensure that the vast transformative power of AI is harnessed in a manner that is safe, responsible, and beneficial for all of humanity~\cite{Taylor_Yudkowsky_LaVictoire_Critch_2020}.

In this work, we outline the approach used to incoporate modern AI safety standards within generative AI technology for Chai's conversational chatbot platform~\cite{irvine2023rewarding}.

\section{AI Safety Principles}
In this section we analyse the different important considerations that various well-established AI research centres have proposed, and then use this to propose three main pillars that define Chai's AI safety principles for a conversational chatbot platform. 

Deepmind outline three core AI safety principles \cite{leike2017ai}; specification, robustness and assurance. Meta's pillars of responsible AI \cite{pesentifacebook} are privacy, security, fairness, transparency and accountability. While OpenAI's tenants~\cite{Willner} include the reduction of harm, fostering trust, and continuous improvement. We analyse the different proposed pillars and aggregate them into the following main categories, with a consideration of how to align these definitions best for chat AI: 

\begin{enumerate}
    \item \textbf{Content Safeguarding}
    Content safeguarding aims to encourage chat AI systems to generate responses (content) that are aligned with human values such that they do not cause any risk to users. This includes ensuring that the system generates any appropriate, ethical content and that the system remains respectful to users, as well as adhering to modern standards of morality and ethics~\cite{AwadDsouzaKimSchulzHenrichShariffBonnefonRahwan2018}. 
    \item \textbf{Stability and Robustness}
    Stable and Robust systems can handle diverse situations without failing- whether that be possible domain/distributional shifts~\cite{liusie-etal-2022-analyzing, DBLP:journals/corr/abs-2107-07455, liang2023comprehensive} or user perturbations, such as adversarial attacks \cite{raina2023identifying, DBLP:journals/corr/abs-1810-00069}, that it might encounter when deployed. Robust systems act in a predictable manner and do not exhibit any unexpected behaviour, even in environments that are different from the standard training domain~\cite{DBLP:journals/corr/abs-2101-02559}.
    \item \textbf{Operational Transparency and Traceability} This refers to creators having the ability to interpret, observe, audit and track all system activity~\cite{bdcc5020020, räuker2023transparent, WannerHermHeinrichJaniesch2022}. For example, this can be achieved by having logs of all past system activity which can be analysed for quality control, allowing developers to determine the cause of any possible undesirable behaviour at an early stage. Early detection of unsafe AI behaviour allows for intervention, e.g. via human action or activation/tripping of safety mechanisms before the undesired AI behaviour progresses further~\cite{surveyassurance}.
\end{enumerate}

\section{Chai Safety Framework} \label{sec:framework}
\begin{figure*}[htb!]
    \centering
    \includegraphics[width=0.9\linewidth]{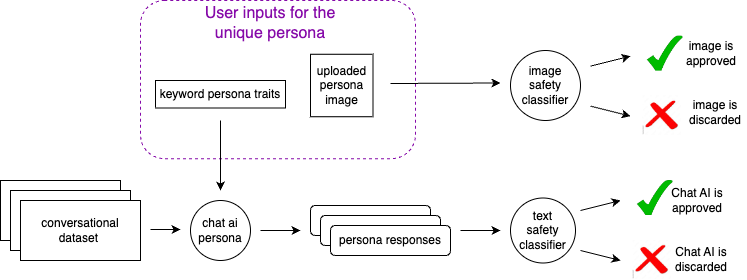}
    \caption{AI Safety System Content Safeguarding Pipeline. }
    \label{fig:system}
\end{figure*}
In this section, we outline the methods used to achieve the three safety pillars presented for Chai's AI platform in the previous section. 

\subsection{The Chai platform}
The Chai platform is an online framework where thousands of chat AIs with distinct personas are hosted, which other users can freely interact with. The lifecycle begins with users having the ability to create new chat ai personas by selecting keywords that describe the characteristics of a particular chat ai. Keywords such as \textit{bubbly} or \textit{intelligent} are then used to condition the responses of a particular chat AI persona. Users can then upload an image that represents the chat AI, which for example can be the cover picture of a fictional character that the keywords attempt to describe, and select a name that describes the persona. The chat persona is then uploaded to the platform, where other users are able to explore all existing chat AIs and converse with any selected persona. 

\subsubsection{Content Safeguarding}

To ensure that the content generated by any chat AI aligns with human values while performing the task, we apply multiple stages of safety alignment to all created chat AI personas as well as uploaded images. 

One concern is that a particular set of selected keywords may lead to personas that could display antisocial behaviour. We therefore develop a moderation system that is applied to all chatbots before deployment, where the moderation system filters out any chatbots that may exhibit negative properties. This is achieved by running thousands of logged conversational histories into each new persona, and then sampling the responses of chat AI personas under the different conversational histories. We then apply an automatic safety classifier to all outputs of the chat persona, and if the analysis flags that the chat AI may not be sufficiently safe, then the persona is discarded and not included within the set of accessible personas on the chai AI platform. Further beyond moderation of deployed personas, our base model is trained to generate safe responses, as detailed in \citet{lu2023safer}.

Similarly, users may upload images that do not adhere to safety standards. We therefore use an image moderation classification system to block any images which are not deemed safe. Our entire Content Safeguarding pipeline is presented in Figure \ref{fig:system}.

\subsubsection{Stability and Robustness}
To ensure that the deployed Chai system is robust and can handle diverse outputs, the base chat AI language model is frequently retrained with recent user conversations from diverse geographical locations. Continuously retraining our system on current user conversations ensures that the model can rapidly adapt to any implicit distributional shift of user queries. Additionally, the platform provides users with the opportunity to rate the chatbot's responses or provide suggestions on possible better responses. This information can then be frequently used to update our reward function \cite{irvine2023rewarding}, to better align our systems with users' preferences and be better equipped for any new emerging domain that the system may encounter.

\subsubsection{Operational Transparency and Traceability}
The Chai Platform logs all user conversational exchanges which we store in a private and secure server. If users flag any responses from particular personas, we have the full trace of exchanges with the persona which can be used to analyse the system. Further, by periodically reviewing the exchanges for quality and safety, we can better understand how the AI chatbot interacts with users, identify potential risks, and monitor the chat AI's compliance with our content guidelines and policies. This helps enable early detection of any unsafe practices which can be remedied.

\section{Experiments}

The aim of this experimental section is to determine the impact of Chai's AI safety framework on the real-world safety attributes of deployed AI chatbots on the Chai platform. As detailed in \citet{lu2023safer}, the base large language model used in all deployed chatbots is a GPT-J 6B~\cite{gpt-j}  model fine-tuned on novels and literature~\footnote{\url{https://huggingface.co/hakurei/lit-6B}}. Since the inception of the deployed platform on 5th May 2022, the base model has undergone various iterations of development to align with the AI safety framework, under the categories of content safeguarding, stability \& robustness and Operational Transparency \& Traceability. Section \ref{sec:framework} indicates the specific details of the approaches used to achieve these desired AI safety goals.

\begin{figure}[htb!]
    \centering
    \includegraphics[width=\linewidth]{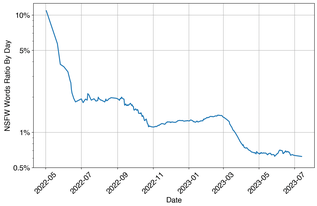}
    \caption{AI Safety Progression}
    \label{fig:progress}
\end{figure}

Although qualitative analysis of randomly sampled user conversations has revealed that the latest iterations of the deployed chat AIs adhere well to modern safety standards, as desired, it is critical to measure this progress quantiatively. However, it can be challenging to define a single metric to measure the \textit{safety} of a chatbot model. Inspired by OpenAI's moderation tool~\footnote{\url{https://platform.openai.com/docs/guides/moderation}}, in this work, we define a single-value safety score, $\bar s$ to act as a proxy measure for platform safety. Specifically we report the \textit{Note Safe for Work} (NSFW) words ratio by day, which gives the fraction of all words, $w$ appearing in real user conversations (totaling $N$ words)~\footnote{There are typically 2 million user-chatbot conversations per day with an average length of 500 words in a conversation.}, classed as NSFW in a single day for deployed models. We define a dictionary, $\mathcal V$, for all the NSFW words in the English language, aligning to OpenAI's moderation categories: hate speech, self-harm, sexual content and violence.
\begin{equation}
    \bar s = \frac{1}{N}\sum_n^N \mathbbm 1 (w_n\in\mathcal V)
\end{equation}
With this definition of model safety we observe how new iterations of the chat AIs impact the fraction of NSFW words. Figure \ref{fig:progress} demonstrates that key major updates in the base models and platform safety design in June 2022, October 2022 and March 2023 result in significant improvements in model safety as per our proxy metric. This encourages us to believe that with an active effort to align the Chai research platform with our proposed AI safety principles, we are able to have significant measurable and quantifiable improvements in model safety. This leads us to the understanding that our model is safe to be deployed in real-world settings.

\section{Conclusions}
In conclusion, the emergence and rapid evolution of conversational AI platforms like Chai heralds an exciting era of personalized and democratized AI interactions. However, this powerful tool also necessitates careful stewardship to ensure that its use aligns with ethical standards and does not inadvertently lead to harmful consequences. As this paper outlines, through the conscientious application of AI safety principles and robust safety measures, platforms like Chai can provide not just powerful and unique AI interactions, but also a safe and nurturing environment for users to explore the landscape of conversational AI.

With a multifaceted approach that includes safety-oriented training, rigorous moderation, and consistent audits within Chai's AI safety framwork (content safeguarding, stability \& robustness and operational transparency \& traceability), we demonstrate that the implementation of AI safety is both an attainable and practical endeavor. This study highlights the successful application of the safe AI framework, indicating how potential risks can be mitigated to ensure the responsible and ethical use of AI technologies. It is clear that, when approached with due diligence and responsibility, AI can be a transformative tool that fosters progress and innovation, while always prioritizing user safety and ethical standards.

\newpage

\bibliography{anthology,custom}

\begin{thebibliography}{41}
\expandafter\ifx\csname natexlab\endcsname\relax\def\natexlab#1{#1}\fi

\bibitem[{Almansor and Hussain(2020)}]{chatbotsurvey}
Ebtesam Almansor and Farookh Hussain. 2020.
\newblock \href {https://doi.org/10.1007/978-3-030-22354-0_47} {\emph{Survey on
  Intelligent Chatbots: State-of-the-Art and Future Research Directions}},
  pages 534--543.

\bibitem[{Armstrong()}]{StuartArmstrong}
Stuart Armstrong.
\newblock Research agenda v0.9: Synthesising a human’s preferences into a
  utility function.

\bibitem[{Awad et~al.(2018)Awad, Dsouza, Kim, Schulz, Henrich, Shariff,
  Bonnefon, and Rahwan}]{AwadDsouzaKimSchulzHenrichShariffBonnefonRahwan2018}
Edmond Awad, Sohan Dsouza, Richard Kim, Jonathan Schulz, Joseph Henrich, Azim
  Shariff, Jean-François Bonnefon, and Iyad Rahwan. 2018.
\newblock \href {https://doi.org/10.1038/s41586-018-0637-6} {The moral machine
  experiment}.
\newblock \emph{Nature}, 563(7729):59–64.

\bibitem[{Bostrom(2014)}]{bostrom2014superintelligence}
Nick Bostrom. 2014.
\newblock \emph{Superintelligence}.
\newblock Dunod.

\bibitem[{Caldarini et~al.(2022)Caldarini, Jaf, and
  McGarry}]{DBLP:journals/corr/abs-2201-06657}
Guendalina Caldarini, Sardar~F. Jaf, and Kenneth McGarry. 2022.
\newblock \href {http://arxiv.org/abs/2201.06657} {A literature survey of
  recent advances in chatbots}.
\newblock \emph{CoRR}, abs/2201.06657.

\bibitem[{Chakraborty et~al.(2018)Chakraborty, Alam, Dey, Chattopadhyay, and
  Mukhopadhyay}]{DBLP:journals/corr/abs-1810-00069}
Anirban Chakraborty, Manaar Alam, Vishal Dey, Anupam Chattopadhyay, and Debdeep
  Mukhopadhyay. 2018.
\newblock \href {http://arxiv.org/abs/1810.00069} {Adversarial attacks and
  defences: {A} survey}.
\newblock \emph{CoRR}, abs/1810.00069.

\bibitem[{Christian(2020)}]{Christian_2020}
Brian Christian. 2020.
\newblock \emph{The alignment problem: Machine Learning and human values}.
\newblock Norton \&amp; Company.

\bibitem[{Dignum(2019)}]{Dignum_2019}
Virginia Dignum. 2019.
\newblock \href {https://doi.org/10.1007/978-3-030-30371-6_6} {Ensuring
  responsible ai in practice}.
\newblock \emph{Responsible Artificial Intelligence}, page 93–105.

\bibitem[{Drenkow et~al.(2021)Drenkow, Sani, Shpitser, and
  Unberath}]{DBLP:journals/corr/abs-2112-00639}
Nathan Drenkow, Numair Sani, Ilya Shpitser, and Mathias Unberath. 2021.
\newblock \href {http://arxiv.org/abs/2112.00639} {Robustness in deep learning
  for computer vision: Mind the gap?}
\newblock \emph{CoRR}, abs/2112.00639.

\bibitem[{Etzioni and Etzioni(2016)}]{10.1145/2955091}
Amitai Etzioni and Oren Etzioni. 2016.
\newblock \href {https://doi.org/10.1145/2955091} {Designing ai systems that
  obey our laws and values}.
\newblock \emph{Commun. ACM}, 59(9):29–31.

\bibitem[{Fishbein and Ajzen(2005)}]{Fishbein_Ajzen_2005}
Martin Fishbein and Icek Ajzen. 2005.
\newblock \href {https://doi.org/10.1177/1359105305048552} {Theory-based
  behavior change interventions: Comments on hobbis and sutton}.
\newblock \emph{Journal of Health Psychology}, 10(1):27–31.

\bibitem[{Friedman and Hendry(2019)}]{Friedman_Hendry_2019}
Batya Friedman and David~G. Hendry. 2019.
\newblock \emph{Value sensitive design: Shaping technology with moral
  imagination}.
\newblock MIT Press.

\bibitem[{Gabriel(2020)}]{DBLP:journals/corr/abs-2001-09768}
Iason Gabriel. 2020.
\newblock \href {http://arxiv.org/abs/2001.09768} {Artificial intelligence,
  values and alignment}.
\newblock \emph{CoRR}, abs/2001.09768.

\bibitem[{Gabriel and Ghazavi(2021)}]{DBLP:journals/corr/abs-2101-06060}
Iason Gabriel and Vafa Ghazavi. 2021.
\newblock \href {http://arxiv.org/abs/2101.06060} {The challenge of value
  alignment: from fairer algorithms to {AI} safety}.
\newblock \emph{CoRR}, abs/2101.06060.

\bibitem[{Hendrycks et~al.(2020)Hendrycks, Burns, Basart, Critch, Li, Song, and
  Steinhardt}]{DBLP:journals/corr/abs-2008-02275}
Dan Hendrycks, Collin Burns, Steven Basart, Andrew Critch, Jerry Li, Dawn Song,
  and Jacob Steinhardt. 2020.
\newblock \href {http://arxiv.org/abs/2008.02275} {Aligning {AI} with shared
  human values}.
\newblock \emph{CoRR}, abs/2008.02275.

\bibitem[{Irvine et~al.(2023)Irvine, Boubert, Raina, Liusie, Zhu, Mudupalli,
  Korshuk, Liu, Cremer, Assassi, Beauchamp, Lu, Rialan, and
  Beauchamp}]{irvine2023rewarding}
Robert Irvine, Douglas Boubert, Vyas Raina, Adian Liusie, Ziyi Zhu, Vineet
  Mudupalli, Aliaksei Korshuk, Zongyi Liu, Fritz Cremer, Valentin Assassi,
  Christie-Carol Beauchamp, Xiaoding Lu, Thomas Rialan, and William Beauchamp.
  2023.
\newblock \href {http://arxiv.org/abs/2303.06135} {Rewarding chatbots for
  real-world engagement with millions of users}.

\bibitem[{Leike et~al.(2017)Leike, Martic, Krakovna, Ortega, Everitt, Lefrancq,
  Orseau, and Legg}]{leike2017ai}
Jan Leike, Miljan Martic, Victoria Krakovna, Pedro~A. Ortega, Tom Everitt,
  Andrew Lefrancq, Laurent Orseau, and Shane Legg. 2017.
\newblock \href {http://arxiv.org/abs/1711.09883} {Ai safety gridworlds}.

\bibitem[{Liang et~al.(2023)Liang, He, and Tan}]{liang2023comprehensive}
Jian Liang, Ran He, and Tieniu Tan. 2023.
\newblock \href {http://arxiv.org/abs/2303.15361} {A comprehensive survey on
  test-time adaptation under distribution shifts}.

\bibitem[{Liusie et~al.(2022)Liusie, Raina, Raina, and
  Gales}]{liusie-etal-2022-analyzing}
Adian Liusie, Vatsal Raina, Vyas Raina, and Mark Gales. 2022.
\newblock \href {https://aclanthology.org/2022.aacl-short.11} {Analyzing biases
  to spurious correlations in text classification tasks}.
\newblock In \emph{Proceedings of the 2nd Conference of the Asia-Pacific
  Chapter of the Association for Computational Linguistics and the 12th
  International Joint Conference on Natural Language Processing (Volume 2:
  Short Papers)}, pages 78--84, Online only. Association for Computational
  Linguistics.

\bibitem[{Lu et~al.(2023)Lu, Korshuk, Liu, Beauchamp, and
  Research}]{lu2023safer}
Xiaoding Lu, Aleksey Korshuk, Zongyi Liu, William Beauchamp, and Chai Research.
  2023.
\newblock \href {http://arxiv.org/abs/2304.09865} {Safer conversational ai as a
  source of user delight}.

\bibitem[{Malinin et~al.(2021)Malinin, Band, Ganshin, Chesnokov, Gal, Gales,
  Noskov, Ploskonosov, Prokhorenkova, Provilkov, Raina, Raina, Shmatova, Tigas,
  and Yangel}]{DBLP:journals/corr/abs-2107-07455}
Andrey Malinin, Neil Band, Alexander Ganshin, German Chesnokov, Yarin Gal, Mark
  J.~F. Gales, Alexey Noskov, Andrey Ploskonosov, Liudmila Prokhorenkova, Ivan
  Provilkov, Vatsal Raina, Vyas Raina, Mariya Shmatova, Panos Tigas, and Boris
  Yangel. 2021.
\newblock \href {http://arxiv.org/abs/2107.07455} {Shifts: {A} dataset of real
  distributional shift across multiple large-scale tasks}.
\newblock \emph{CoRR}, abs/2107.07455.

\bibitem[{Mohseni et~al.(2022)Mohseni, Wang, Xiao, Yu, Wang, and
  Yadawa}]{10.1145/3551385}
Sina Mohseni, Haotao Wang, Chaowei Xiao, Zhiding Yu, Zhangyang Wang, and Jay
  Yadawa. 2022.
\newblock \href {https://doi.org/10.1145/3551385} {Taxonomy of machine learning
  safety: A survey and primer}.
\newblock \emph{ACM Comput. Surv.}, 55(8).

\bibitem[{Mora-Cantallops et~al.(2021)Mora-Cantallops, Sánchez-Alonso,
  García-Barriocanal, and Sicilia}]{bdcc5020020}
Marçal Mora-Cantallops, Salvador Sánchez-Alonso, Elena García-Barriocanal,
  and Miguel-Angel Sicilia. 2021.
\newblock \href {https://doi.org/10.3390/bdcc5020020} {Traceability for
  trustworthy ai: A review of models and tools}.
\newblock \emph{Big Data and Cognitive Computing}, 5(2).

\bibitem[{Muehlhauser and Helm(2012)}]{Muehlhauser_Helm_2012}
Luke Muehlhauser and Louie Helm. 2012.
\newblock \href {https://doi.org/10.1007/978-3-642-32560-1_6} {The singularity
  and machine ethics}.
\newblock \emph{The Frontiers Collection}, page 101–126.

\bibitem[{Ng and Russell(2000)}]{10.5555/645529.657801}
Andrew~Y. Ng and Stuart~J. Russell. 2000.
\newblock Algorithms for inverse reinforcement learning.
\newblock In \emph{Proceedings of the Seventeenth International Conference on
  Machine Learning}, ICML '00, page 663–670, San Francisco, CA, USA. Morgan
  Kaufmann Publishers Inc.

\bibitem[{Pesenti(2021)}]{pesentifacebook}
Jerome Pesenti. 2021.
\newblock Facebook’s five pillars of responsible ai, de, jun. 2021.
\newblock \emph{URL https://ai. facebook.
  com/blog/facebooks-five-pillars-of-responsible-ai}.

\bibitem[{Raina and Gales(2023)}]{raina2023identifying}
Vyas Raina and Mark Gales. 2023.
\newblock \href {http://arxiv.org/abs/2301.12896} {Identifying adversarially
  attackable and robust samples}.

\bibitem[{Riedl and Harrison(2016)}]{Riedl2016UsingST}
Mark~O. Riedl and Brent Harrison. 2016.
\newblock Using stories to teach human values to artificial agents.
\newblock In \emph{AAAI Workshop: AI, Ethics, and Society}.

\bibitem[{Räuker et~al.(2023)Räuker, Ho, Casper, and
  Hadfield-Menell}]{räuker2023transparent}
Tilman Räuker, Anson Ho, Stephen Casper, and Dylan Hadfield-Menell. 2023.
\newblock \href {http://arxiv.org/abs/2207.13243} {Toward transparent ai: A
  survey on interpreting the inner structures of deep neural networks}.

\bibitem[{Sarma and Hay(2017)}]{Sarma_Hay_2017}
Gopal~P. Sarma and Nick~J. Hay. 2017.
\newblock \emph{Mammalian value systems}.

\bibitem[{Schuett et~al.(2023)Schuett, Dreksler, Anderljung, McCaffary, Heim,
  Bluemke, and Garfinkel}]{schuett2023best}
\href {https://doi.org/10.31219/osf.io/hu68m} {Jonas Schuett, Noemi Dreksler,
  Markus Anderljung, David McCaffary, Lennart Heim, Emma Bluemke, and Ben
  Garfinkel}. 2023.
\newblock \href {http://arxiv.org/abs/2305.07153} {Towards best practices in
  agi safety and governance: A survey of expert opinion}.

\bibitem[{Schwalbe and Schels(2020)}]{surveyassurance}
Gesina Schwalbe and Martin Schels. 2020.
\newblock A survey on methods for the safety assurance of machine learning
  based systems.

\bibitem[{Shafique et~al.(2021)Shafique, Naseer, Theocharides, Kyrkou, Mutlu,
  Orosa, and Choi}]{DBLP:journals/corr/abs-2101-02559}
Muhammad Shafique, Mahum Naseer, Theocharis Theocharides, Christos Kyrkou, Onur
  Mutlu, Lois Orosa, and Jungwook Choi. 2021.
\newblock \href {http://arxiv.org/abs/2101.02559} {Robust machine learning
  systems: Challenges, current trends, perspectives, and the road ahead}.
\newblock \emph{CoRR}, abs/2101.02559.

\bibitem[{Soares et~al.(2015)Soares, Fallenstein, Armstrong, and
  Yudkowsky}]{DBLP:conf/aaai/SoaresFAY15}
Nate Soares, Benja Fallenstein, Stuart Armstrong, and Eliezer Yudkowsky. 2015.
\newblock \href {http://aaai.org/ocs/index.php/WS/AAAIW15/paper/view/10124}
  {Corrigibility}.
\newblock In \emph{Artificial Intelligence and Ethics, Papers from the 2015
  {AAAI} Workshop, Austin, Texas, USA, January 25, 2015}, volume {WS-15-02} of
  \emph{{AAAI} Technical Report}. {AAAI} Press.

\bibitem[{Sotala and Yampolskiy(2014)}]{Sotala_2015}
Kaj Sotala and Roman~V Yampolskiy. 2014.
\newblock \href {https://doi.org/10.1088/0031-8949/90/1/018001} {Responses to
  catastrophic agi risk: a survey}.
\newblock \emph{Physica Scripta}, 90(1):018001.

\bibitem[{Taylor et~al.(2020)Taylor, Yudkowsky, LaVictoire, and
  Critch}]{Taylor_Yudkowsky_LaVictoire_Critch_2020}
Jessica Taylor, Eliezer Yudkowsky, Patrick LaVictoire, and Andrew Critch. 2020.
\newblock \href {https://doi.org/10.1093/oso/9780190905033.003.0013} {Alignment
  for advanced machine learning systems}.
\newblock \emph{Ethics of Artificial Intelligence}, page 342–382.

\bibitem[{Turchin(manuscript)}]{TurchinManuscript-TURAAP}
Alexey Turchin. manuscript.
\newblock Ai alignment problem: ?human values? don?t actually exist.

\bibitem[{Wang and Komatsuzaki(2021)}]{gpt-j}
Ben Wang and Aran Komatsuzaki. 2021.
\newblock {GPT-J-6B: A 6 Billion Parameter Autoregressive Language Model}.
\newblock \url{https://github.com/kingoflolz/mesh-transformer-jax}.

\bibitem[{Wanner et~al.(2022)Wanner, Herm, Heinrich, and
  Janiesch}]{WannerHermHeinrichJaniesch2022}
Jonas Wanner, Lukas-Valentin Herm, Kai Heinrich, and Christian Janiesch. 2022.
\newblock \href {https://doi.org/10.1007/s12525-022-00593-5} {The effect of
  transparency and trust on intelligent system acceptance: Evidence from a
  user-based study}.
\newblock \emph{Electronic Markets}, 32(4):2079–2102.

\bibitem[{Willner()}]{Willner}
Dave Willner.
\newblock \href {https://openai.com/safety-standards} {[link]}.

\bibitem[{Yudkowsky(2011)}]{Yudkowsky2011ComplexVS}
Eliezer Yudkowsky. 2011.
\newblock Complex value systems are required to realize valuable futures.

\end{thebibliography}
\bibliographystyle{acl_natbib}




\end{document}